\documentclass[letterpaper, 10 pt, journal, twoside]{IEEEtran}

\ifCLASSINFOpdf
\else
\fi

\usepackage{multicol}
\usepackage{multirow}
\usepackage[utf8]{inputenc} 
\usepackage[T1]{fontenc}    
\usepackage{url}            
\usepackage{booktabs}       
\usepackage{amsfonts}       
\usepackage{nicefrac}       
\usepackage{microtype}      
\usepackage{amsmath,amssymb}
\usepackage{mathtools}
\usepackage{array}
\usepackage{algpseudocode}
\usepackage{algorithm}
\usepackage{accents}
\usepackage{tabstackengine}
\usepackage[table]{xcolor} 
\algtext*{EndWhile} 
\algtext*{EndIf} 
\algtext*{EndProcedure}
\algtext*{EndFor} 
\usepackage{siunitx} 
\usepackage{lipsum}
\usepackage{float}  
\usepackage{wrapfig}
\usepackage{subcaption}
\usepackage{cite}

\usepackage{caption} 
\usepackage{pifont} 
\definecolor{lightgray}{gray}{0.93} 
\definecolor{bclone}{HTML}{E4F0F6}
\definecolor{dagger}{HTML}{E8F6DC}
\definecolor{ibc}{HTML}{FAECD5}
\definecolor{gail}{HTML}{FDE0E0}
\definecolor{act}{HTML}{cea39f}
\definecolor{diffusion}{HTML}{d9c1e7}

\newcommand{\bm}[1]{\boldsymbol{\mathrm{#1}}}

\begin{document}
%
\title{A Comparison of Imitation Learning Algorithms for Bimanual Manipulation}
%
%
%

\author{Michael Drolet$^{1,4}$, Simon Stepputtis$^{2*}$, Siva Kailas$^{2*}$, Ajinkya Jain$^{3}$, \\ Jan Peters$^{4}$, Stefan Schaal$^{3}$, and Heni Ben Amor$^{1}$

\thanks{$^{1}$Michael Drolet and Heni Ben~Amor are with the School of Computing and Augmented Intelligence at Arizona State University, Tempe, USA {\tt\footnotesize \{mdrolet, hbenamor\}@asu.edu}}%
\thanks{$^{2}$Simon Stepputtis and Siva Kailas are with the Robotics Institute at Carnegie Mellon University, Pittsburgh, USA {\tt\footnotesize \{sstepput, skailas\}@andrew.cmu.edu}, * Equal contribution}
\thanks{$^{3}$Ajinkya Jain and Stefan Schaal are with [Google] Intrinsic, Mountain View, USA {\tt\footnotesize \{ajinkyajain, sschaal\}@google.com}}
\thanks{$^{4}$Jan Peters and Michael Drolet are with the Department of Computer Science at TU Darmstadt, Darmstadt, Germany {\tt\footnotesize \{michael.drolet,jan.peters\}@tu-darmstadt.de}}
}

%
%

\markboth{IEEE Robotics and Automation Letters. Preprint Version}
{Drolet \MakeLowercase{\textit{et al.}}: A Comparison of Imitation Learning Algorithms for Bimanual Manipulation} 

%



\maketitle

\begin{abstract}
Amidst the wide popularity of imitation learning algorithms in robotics, their properties regarding hyperparameter sensitivity, ease of training, data efficiency, and performance have not been well-studied in high-precision industry-inspired environments.
In this work, we demonstrate the limitations and benefits of prominent imitation learning approaches and analyze their capabilities regarding these properties.
We evaluate each algorithm on a complex bimanual manipulation task involving an over-constrained dynamics system in a setting involving multiple contacts between the manipulated object and the environment. 
While we find that imitation learning is well suited to solve such complex tasks, not all algorithms are equal in terms of handling environmental and hyperparameter perturbations, training requirements, performance, and ease of use.
We investigate the empirical influence of these key characteristics by employing a carefully designed experimental procedure and learning environment.
\end{abstract}


%
\IEEEpeerreviewmaketitle

\section{Introduction}
\label{sec:introduction}

Bimanual manipulation is a critical motor skill that has allowed humans and primates to create tools and use them successfully. From handling stone tools for food processing to modern day engineering and repair of intricate machines, bimanual manipulation has played an important role in the development of mankind.
Although mundane tasks such as tying knots and unboxing items may seem straightforward to most adults, they pose significant challenges for robots due to difficulties in perception, planning, and control, especially in contact-rich dexterous manipulation.

Bimanual manipulation also allows a robot to simultaneously hold multiple objects -- one in each hand -- a feat unattainable for a single-arm robot. 
Additionally, using both arms concurrently enables the robot to expedite task completion and handle heavier objects by selecting grasps that distribute the load more effectively \cite{stepputtis2022system, kruger2011dual}. 
Thus, bimanual manipulation offers a promising pathway to equip robots with sophisticated motor skills comparable to humans.

\begin{figure}[t]
    \centering
    \includegraphics[width=0.485\textwidth]{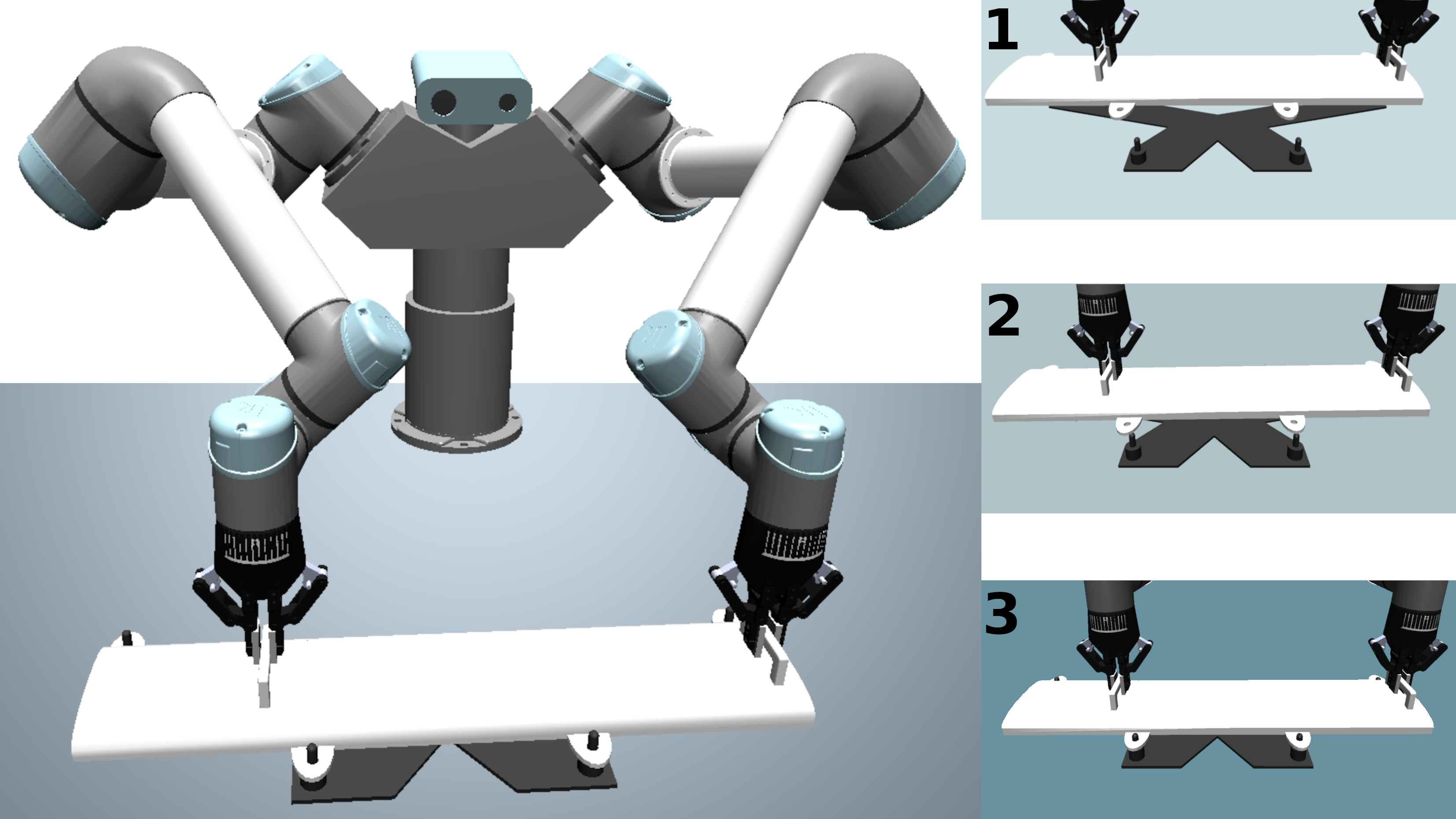}
    \caption{Two UR5 arms equipped with grippers and mounted to a rotating torso. The robot above completes the final stage of the high-precision four-peg insertion task.}
    \label{fig:insertion_sequence}
\end{figure}

Learning bimanual manipulation skills on a robot can be approached in several ways, and we briefly discuss two of the prominent approaches.
One possible strategy is through reinforcement learning (RL). 
RL enables an agent to iteratively interact with its environment, developing a control policy based on the information/rewards provided by the environment \cite{sutton2018reinforcement}.
Such an approach can be advantageous, considering the potential to discover novel strategies. 
However, applying RL to tasks in the real world can lead to undesirable policies \cite{randlov1998learning}: 
Without properly designed reward functions \cite{eschmann2021reward}, the agent is amenable to exploiting niches of the environment that create inappropriate, non-robust, and potentially unsafe solutions.
Designing a reward function that accurately captures the desired behavior and encourages the robot to achieve the task goals is often non-trivial. 
Imitation learning (IL, also called behavioral cloning (BC), or learning from demonstration (LfD)), on the other hand, does not require a reward function to be explicitly defined. 
Many IL algorithms such as DMPs \cite{schaal2005learning} are computationally efficient (and sometimes used in a one-shot manner), greatly reducing the risk of hardware degradation.
This efficiency is particularly valuable in real-world scenarios where extensive interaction with the environment is impractical. The drawback of IL is that it cannot discover new solutions outside the distribution of the training data, but adding new data to cover problems is often straightforward.
Motivated by these advantages, and also the fact that IL has gained a lot of popularity in recent years, we focus on IL approaches in this work and benchmark several IL algorithms for performing a high-precision bimanual peg insertion task from the practical, computational data efficiency, and performance perspectives.

Given IL's desirable properties and the importance of bimanual manipulation in achieving more sophisticated human-like robots, a natural question is how to approach the intersection of these two fields. 
We address this question directly by combining 1) several foundational algorithms in IL with 2) a benchmark MuJoCo \cite{todorov2012mujoco} environment that seeks to fairly and extensively compare algorithms in terms of sample efficiency, noise robustness, compute time, and performance. 
In doing so, we provide an extensive discussion related to the various advantages and disadvantages of these algorithms as well as the engineering approaches that allow for learning in such an environment.
More specifically, we focus on Generative Adversarial Imitation Learning (GAIL) \cite{ho2016generative}, Implicit Behavioral Cloning (IBC) \cite{florence2022implicit}, Dataset Aggregation (DAgger) \cite{ross2011reduction}, Behavioral Cloning (BC) \cite{pomerleau1988alvinn}, Acting Chunking Transformer (ACT) \cite{zhao2023learning}, and the Diffusion Policy \cite{chi2023diffusion}. 
In the environment, the robot learns to transfer an adapter with four holes and insert it into a stationary adapter with four pegs. 
The difficulty in the task lies within the low tolerance of the adapters, such that success only occurs when the robot is precise;  i.e., the holes have a diameter of 11mm and the rounded pins have a base diameter of 10mm, leaving approximately 1mm of tolerance.

\section{Related Work}
Imitation learning is a paradigm that enables agents to learn directly from experts in the form of demonstrations \cite{fang2019survey,hussein2017imitation, argall2009survey}. 
In particular, imitation learning has been a useful strategy for learning various robotic tasks \cite{schaal2005learning, chalodhorn2007learning, learningMotorLfD}, especially manipulation-based tasks such as pick-and-place or grasping \cite{GMMLfD, billard2004discovering, mulling2013learning}.
Additionally, research in bimanual manipulation is still emerging in robotics \cite{krebs2022bimanual}. 
Early work in bimanual manipulation utilized classical control-based approaches \cite{mirrazavi2018unified}, and a survey of existing works that primarily utilized control-based approaches with known environment dynamics was presented in \cite{SMITH20121340}. 
Model-based approaches that utilize planning and constraint solving have also been used to perform cloth folding and scooping \cite{6095109,grannen2023learning}. 

\begin{figure}[t]
    \centering
    \vspace{2mm}
    \includegraphics[width=0.49\textwidth]{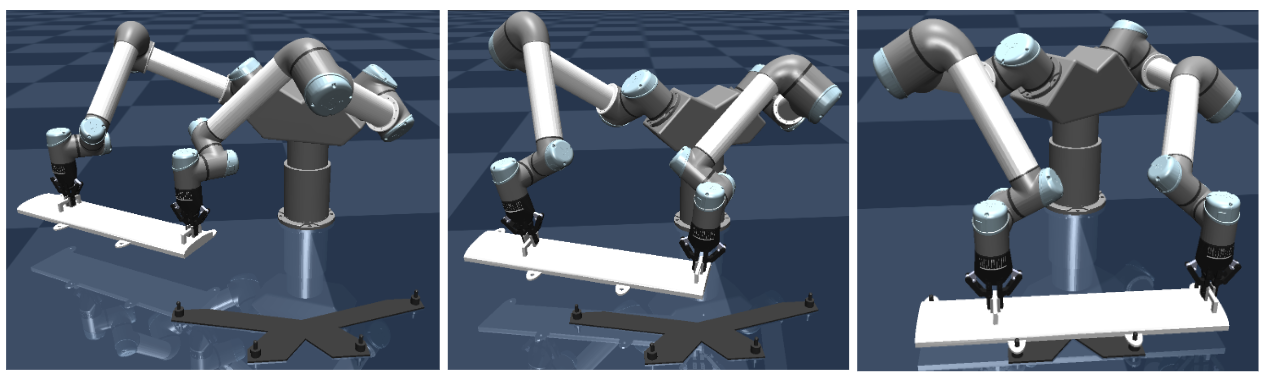}
    \caption{The robot successfully transfers and inserts the dynamic adapter (white) into the stationary adapter (black).}
    \label{fig:insertion_sequence}
\end{figure}

More recently, learning-based approaches have emerged in the bimanual manipulation domain. 
Reinforcement learning is one class of approaches that have been proposed~\cite{lin2023bi, kataoka2022bi}. 
For example, \cite{lin2023bi} presents a set of bimanual manipulation tasks and associated reward structures that were empirically found to work well with deep reinforcement learning. 
One study also utilized a marker-based vision system, sim-to-real, and reinforcement learning for connecting two blocks with magnetic connection points with two robotic arms \cite{kataoka2022bi}.

A method for mastering contact-rich manipulation in a similar setup has been proposed, using motor primitives to train the robot for an insertion task \cite{stepputtis2022system}. 
This approach involves utilizing force feedback and environment feedback as stimuli during the filtering process outlined in Bayesian Interaction Primitives \cite{campbell2017bayesian}. 
Although this method displays efficacy, our objective is to eliminate the inductive bias associated with motor primitives and explore neural network-based approaches instead. 
In doing so, we adopt a more general class of function approximation.

ALOHA is a recent approach to learning several fine-grained bimanual manipulation tasks with everyday objects \cite{zhao2023learning}. 
This approach is notable due to its high degree of success on many tasks that -- to our knowledge -- were previously only achievable by human demonstrators.
ALOHA presents an action-chunking transformer (ACT), which we implement for comparison in this work. 
Although there are related works in the area of bimanual manipulation, such as HDR-IL \cite{xie2020deep} and SIMPLe \cite{10215052}, we seek to investigate algorithms that have been widely used over the last several years in robotics and require a minimal number of demonstrations.

A separate study discusses the essential components of training adversarial imitation learning algorithms \cite{orsini2021matters}. 
This study is extensive in terms of evaluating different hyperparameters, discriminator configurations, and training-related metrics. 
In this investigation, we are interested in not only adversarial methods but also how non-adversarial methods compare in the context of bimanual manipulation. 
Furthermore, other studies evaluate various imitation learning algorithms and their hyperparameters \cite{arulkumaran2021pragmatic, hussenot2021hyperparameter}. 
However, key algorithms such as DAgger and IBC have been excluded from the scope of these studies. 
Moreover, the gym environments employed in these studies fall short in directly capturing the nuanced dynamics and fine-grained behaviors inherent in our bimanual robot setup.

\section{Algorithm Selection}
The selection of algorithms is a critical point of consideration; consequently, the chosen algorithms can be seen as an orthogonal set of approaches to imitation learning in general. 
On one hand, the offline and supervised learning (SL) aspect of IL is captured by both an expressive energy-based policy implementation (IBC) and the widely familiar Gaussian neural network-based policy (BC). 
On the other hand, methods that interact with the environment in the form of (a) an oracle (DAgger) to minimize covariate shift, and (b) a reinforcement learning policy (GAIL) help capture the class of approaches reliant on sampling states online. 
We additionally adopt some of the most recent and successful methods in IL for robotics, namely ACT and Diffusion Policy.
We believe that while there are many derivatives of these methods, their longstanding impact on the field of imitation learning helps justify the need to compare them.

\subsection{Behaviorial Cloning}
\label{sec:bc}
Behavioral Cloning (BC) is one of the most well-known and widely used imitation learning algorithms, largely due to its simplicity and effectiveness on large datasets. BC is typically implemented using the following objective:
\begin{equation}
     \hat{\bm{\theta}} =  \operatorname*{argmax}_{\bm{\theta}} \mathbb{E}_{(\bm{s},\bm{a}) \sim \bm{\tau}_{E}}\left[\log(\pi_{\bm{\theta}}(\bm{a} | \bm{s})) \right]
\end{equation}
where $\bm{\tau}_{E}$ represents the state-action trajectories from the expert, $\pi_{\bm{\theta}}$ is the current policy, $\bm{a}$ is a continuous action, and $\bm{s}$ is the observed continuous state. We additionally include a tunable $L_1$ and $L_2$ penalty on the parameters of the policy (known as elastic net regularization) to help prevent overfitting. In the context of this paper, we refer to BC as solely training a Gaussian policy whose mean is given by a feed-forward neural network, $\pi(\bm{a} | \bm{s})$. 
Recent successes in behavioral cloning for robot manipulation include the RT-X \cite{padalkar2023open} model and its precursors, all trained on very large datasets (each consisting of $130,000$ training demonstrations or more).
Our work, however, studies BC in the small data regime, using a maximum of $200$ expert demonstrations.

\subsection{Action Chunking Transformer}
The Action Chunking Transformer (ACT) \cite{zhao2023learning} performs behavioral cloning using a conditional variational autoencoder (CVAE) implemented as a multi-headed attention transformer \cite{vaswani2017attention}. 
While the objective for training ACT is largely the same as vanilla BC (Section \ref{sec:bc}), we will briefly describe its differences from the standard formulation. For one, ACT is trained (as in the original work) using the $L_1$ loss, which can be interpreted as being proportional to the square root of the Mahalanobis distance used in the objective of vanilla BC, assuming a fixed variance. This deterministic policy then predicts a sequence of actions instead of a single action and- in our formulation- uses a history of observations as input to the transformer instead of image observations.

\subsection{Implicit Behavorial Cloning}
Implicit Behavioral Cloning (IBC) \cite{florence2022implicit} is a supervised learning approach using Energy-Based Models. 
IBC is trained using the Negative Counter Example (NCE) loss function, such that negative counter-examples of the expert are generated to train the model \cite{oord2018representation}.
In this method, energies are assigned to the state-action pairs, and the policy takes the action that minimizes the energy landscape.
As the minimum over the actions is taken, IBC has the advantage of handling discontinuities that can arise in the typical regression setting, where behavioral cloning may simply interpolate.
This is a desirable feature of implicit models, and it is one of the presented advantages in the IBC work that makes it unique compared to other imitation learning algorithms. In short, the IBC policy can be summarized as:
\begin{equation}
    \hat{\bm{a}} = \operatorname*{argmin}_{\bm{a}} E_{\bm{\theta}}(\bm{s}, \bm{a})
\end{equation}
where $E_{\bm{\theta}}$ is the energy function and $\hat{\bm{a}}$ is the optimal action.
However, many works have found that the IBC objective is numerically unstable and does not consistently yield high-quality policies \cite{singh2023revisiting}. 

\subsection{Diffusion Policy}
Like the implicit policy presented in IBC, the Diffusion Policy \cite{chi2023diffusion} performs an iterative procedure to generate actions. 
Diffusion models have achieved significant success in areas like image generation \cite{ho2020denoising}. 
Their observed stability, compared to energy-based models, makes them a promising method to explore in the robotics domain.
Using a series of denoising steps, this method presents a way to refine noise into actions via a learned gradient field. 
The Diffusion Policy in this work is implemented using a U-Net architecture, which conditions on an observation history and generates an action sequence similar to ACT.
Algorithms such as IBC and Diffusion Policy (based on Langevin dynamics) provide viable alternatives to the standard behavioral cloning formulation, which may lack the expressiveness these models provide.

\subsection{Generative Adversarial Imitation Learning}

GAIL formulates imitation learning as an inverse reinforcement learning (IRL) problem, wherein the reward function is learned based on the discriminator's scores \cite{ho2016generative}. 
The discriminator is a component of the Generative Adversarial Network (GAN) \cite{goodfellow2020generative} setup, such that the generator network (i.e., the policy) tries to produce state-action pairs that match the expert's as closely as possible. 
A suitable policy has been learned once the discriminator can no longer differentiate between state-action samples from the policy and expert (due to the policy's "expert-like behavior"). 
The parameters $\bm{w}$ for the discriminator $D$ are updated using the following objective:
\begin{equation}
    \mathbb{\hat{E}}_{\bm{\tau}_i}\left[\nabla_{\bm{w}} \log\left(D_{\bm{w}}(\bm{s},\bm{a})\right)\right] + \mathbb{\hat{E}}_{\bm{\tau}_E}\left[\nabla_{\bm{w}} \log\left(1 - D_{\bm{w}}(\bm{s},\bm{a})\right)\right]
\end{equation}
where $\bm{\tau}_i$ represents state-action trajectories from the most recent policy (at iteration $i$).
The policy is updated using a standard policy gradient algorithm such as, in our case, Trust Region Policy Optimization (TRPO) \cite{schulman2015trust}.

Methods such as VAIL \cite{peng2018variational} seek to address the issue of generator/discriminator imbalance by using a variational bottleneck to constrain the gradient updates of the networks.
While there are many derivatives of GAIL \cite{li2017infogail, xiao2019wasserstein}, we believe the analysis is best done by using the original motivating work, and hence we use the original implementation for our study. 

\begin{table}[t]
    \vspace{2mm}
    \setlength{\tabcolsep}{6pt} 
    \renewcommand{\arraystretch}{1} 
    \resizebox{0.485\textwidth}{!}{%
        \begin{tabular}{l|c|c|c}
            \hline
            \rowcolor{lightgray}
            \textbf{Algorithm} & \textbf{Env. Interaction} & \textbf{Policy Class} & \textbf{Train $\pi$} \\
            \hline
            BC, ACT & \ding{55} & Gaussian, Deterministic & SL \\
            IBC, Diffusion & \ding{55} & EBM, Langevin & SL \\
            DAgger & \ding{51} & Gaussian & SL \\
            GAIL & \ding{51} & Gaussian & RL \\
            \hline
        \end{tabular}%
    }
    \captionsetup{skip=5pt}
    \caption{Comparison of Algorithms.}
    \label{tab:algorithm_comparison}
\end{table}

\subsection{DAgger: Dataset Aggregation}
DAgger addresses the covariate shift problem, where the distribution of observations the policy encounters differs from those in the expert dataset \cite{ross2011reduction}. 
To tackle this challenge, a data aggregation scheme is employed wherein the policy is re-trained on the history of expert-labeled states encountered over time. 
The need to specify an optimal action at all possible states without using a human can also make implementing DAgger challenging.
Additionally, the learner's capacity to make decisions raises safety apprehensions, especially in critical settings such as autonomous driving.

DAgger theoretically bounds the training loss of the best policy under its distribution of sampled trajectories, where the tightness of this bound depends on, e.g., the number of iterations, samples per iteration, and mixing coefficient.
However, its practical efficacy is constrained by factors such as the quality of the oracle and the difficulty of the environment. 
Many approaches have improved upon the fundamental idea of DAgger by using a set of expert demonstrations to improve the explorative learning process~\cite{ kelly2019hg, zhang2016query}. 
However, while such improvements have been created with a focus on safety, multi-agent settings, and a mixture of experts, to this day, DAgger remains a fundamental algorithm worth studying.

\section{Methodology}
In the following section, we describe the methods for creating the bimanual manipulation insertion expert, as well as the design considerations necessary for learning with such a system. 
The implementation contains a two-stage expert, as outlined in Algorithm \ref{alg:cap}. 
A dynamics model is proposed that allows for implicit control of the torso, such that the state formulation (described in Section \ref{sec:environments}) can be predominantly characterized by the end-effectors.
The robot consists of two UR5 arms, each mounted to a rotating torso and equipped with Robotiq 2F-85 grippers.

\subsection{Operational Space Controller (OSC)}
\begin{figure}[t!]
\raggedright
\resizebox{\linewidth}{!}{%
\begin{minipage}{1.0\linewidth}
\begin{algorithm}[H]
\caption{Bimanual Insertion Expert}\label{alg:cap}
\centering
\begin{algorithmic}[1]
\linespread{0.9}\selectfont
\Procedure{pathFollowExpert}{$i, t$}
        \State{$\bm{\rho}, \bm{\phi} =  \Call{getRobotState}$ ; $t' =  \Call{min}{t, T_i}$}
        \For{$j$ in $[1,...,J]$}
            \State{$\Delta \bm{\rho}^j =  \Call{feedbackController}{} (\bm{\rho}^j - \bm{\rho}^{i,j}_{\ast}(t'))$}
            \State{$\Delta \bm{\phi}^j = \Call{clip}{} ( \Call{diff}{} (\bm{\phi}^j , \bm{\phi}^{i,j}_{\ast}(t')))$}
        \EndFor
            \State{\Return $\left([\Delta \bm{\rho}^1,...,\Delta \bm{\rho}^J], g([\Delta \bm{\phi}^1,...,\Delta \bm{\phi}^J])\right)$}
\EndProcedure \\
\vspace{-2mm}
\Procedure{GetExpertAction}{$i, t$}
    \If{\Call{doPathFollow}{}}
    \State{$(\Delta \bm{\rho}, \Delta \bm{\phi}) = \Call{pathFollowExpert}{i, t}$}
    \ElsIf{\Call{doInsertion}{}}
    \State{$(\Delta \bm{\rho}, \Delta \bm{\phi}) = \Call{insertExpert}$ \hspace{-6mm} \Comment{similar to 1:}} 
    \EndIf
    \State{\Return $(\Delta \bm{\rho}, \Delta \bm{\phi})$}
\EndProcedure \\
\vspace{-2mm}
\Procedure{GenerateExpertDemo}{$i$}
    \State{$\mathcal{D}_i \gets \emptyset$ ;  $t = 1$  ; $\Delta \bm{\phi} = \bm{0}$  ;  $\Delta \bm{\rho} = \bm{0}$}
    \While{not \Call{DoneInsert}{}}
        \State{$\bm{\rho}, \bm{\phi} =  \Call{GetRobotState}$}
        \State{$\bm{s} = \Call{ApplyOSC}{\bm{\rho} + \Delta \bm{\rho}, \bm{\phi} + f(\Delta \bm{\phi}})$}
        \State{$\bm{a} = (\Delta \bm{\rho}, \Delta \bm{\phi}) = \Call{GetExpertAction}{i, t}$}
        \State{$\mathcal{D}_i \gets \mathcal{D}_i \cup (\bm{s}, \bm{a})$ ; $t = t + 1$}
    \EndWhile
    \State{\Return $\mathcal{D}_i$}
\EndProcedure
\end{algorithmic}
\label{alg:Expert}
\end{algorithm}
\end{minipage}
}
\vspace{-2mm}
\end{figure}
The operational space controller (OSC) \cite{khatib1987unified} is used in this work to facilitate learning in the task space \cite{peters2008learning}.
Although more task-specific controllers may be suitable for this task, such as the cooperative dual task space representation \cite{adorno2010dual, caccavale2000task, uchiyama1992symmetric}, our analysis is primarily concerned with the choice of learning algorithms (all of which will be similarly affected by the choice of controller), and hence we opt for a more general task space control framework that may be more redundantly specified.
Given a desired 3D position and rotation for each controlled DoF $j \in [1,\cdots,J]$, we denote the task space errors as
$\tilde{\bm{x}}_{\text{task}} = \begin{bmatrix} \tilde{\bm{\rho}}^{1}, \tilde{\bm{\phi}}^{1}, \cdots , \tilde{\bm{\rho}}^{J}, \tilde{\bm{\phi}}^{J} \end{bmatrix}$, where $\bm{\rho}$ and $\bm{\phi}$ are positions and rotations, respectively. 
These values are saturated by maximum velocity constraints and scaled by proportional gains as in \cite{khatib1986real}.
An admittance controller is adopted, wherein both end-effectors have a force/torque sensor providing measurements in the task space. 
The measured forces $\bm{f}_{\text{ext}}$ are utilized to move the robot arms in the same direction.
These forces are scaled by a gain matrix $\bm{K}_x$ such that the final error term is $\tilde{\bm{x}} = \tilde{\bm{x}}_{\text{task}} - \bm{K}_x \bm{f}_{\text{ext}}$.
As the action space does not include a target velocity, we compensate for velocity in the joint space by taking the error from a target joint velocity of $\bm{0}$, i.e., $\tilde{\dot{\bm{q}}} = \bm{K}_v(\bm{0} - \dot{\bm{q}})$.
    
Given that $J = 2$ (one left arm and one right arm) and the rotation errors are expressed in Euler angles, it follows that $\bm{x}$ is $12$-dimensional.
We adopt the Jacobian pseudo-inverse method, using the dynamically consistent generalized inverse, as shown in Equation \ref{eq:jac_update}. 
Because we are controlling two UR5 arms (each with six DoF) and one base joint implicitly, $\bm{q}$ is $13$-dimensional.
Consequently, we have the Jacobian $\bm{J}(\bm{q}) \in \mathbb{R}^{12 \times 13}$, inertia matrix $\bm{M}(\bm{q}) \in \mathbb{R}^{13 \times 13}$, and forces due to gravity $\bm{g}(\bm{q}) \in  \mathbb{R}^{13}$.
The force vector used to control the robot is then calculated as:
\begin{equation}
\bm{u} = \bm{J}(\bm{q})^\top\bm{M}_x(\bm{q})\tilde{\bm{x}} + \bm{M}(\bm{q})\tilde{\dot{\bm{q}}} + \bm{g}(\bm{q})
\label{eq:jac_update}
\end{equation}
where $\bm{M}_x(\bm{q}) = (\bm{J}(\bm{q}) \bm{M}(\bm{q})^{-1} \bm{J}(\bm{q})^\top)^\dagger$. Finally, we apply a nullspace filter to the force output:
\begin{equation}
\bm{u} = \bm{u} + \left(\bm{I} - \bm{J}(\bm{q})^\top \bar{\bm{J}}(\bm{q})^\top\right)\bm{u}_{\text{null}}
\end{equation}
where $\bar{\bm{J}}(\bm{q}) = \bm{M}^{-1}(\bm{q}) \bm{J}(\bm{q})^\top \bm{M}_x(\bm{q})$ and $\bm{u}_{\text{null}} = \bm{K}_n\bm{M}(\bm{q})(\dot{\bm{q}}_* - \dot{\bm{q}})$. We take $\dot{\bm{q}}_*$ to be $\bm{0}$ (as in the velocity controller), and $\bm{K}_n$ is a parameterized diagonal matrix.

\subsection{Bimanual Manipulation Expert Controller}
\label{sec:bm_expert}

We are first given $n$ \textit{original} demonstrations, where the $i$'th demonstration contains positions ${\bm{\rho}}^{i,j}_{\ast}(t) \in \mathbb{R}^{3}$ and quaternions ${\bm{\phi}}_{\ast}^{i,j}(t) \in \bm{so}(3)$ for all timesteps $t \in [1,2,...,T_i]$ for DoF $j$. 
Every original demonstration is then converted to a sequence of expert state-action pairs using the $\Call{generateExpertDemo}{}$ procedure. 
In doing so, we obtain state-action pairs from the same environment and robot used during training. 
A separate representation space (6DRR) is used in this procedure to transform to-and-from our internal representation of quaternions \cite{zhou2019continuity}. 
Here, the co-domain of the forward mapping $g(.)$ represents the continuous representation space in which the neural networks are trained. This function returns the first two columns of a rotation matrix.
The backward mapping $f(.)$ is then used to transform back to a rotation matrix using the Gram-Schmidt process.

Algorithm 1 describes the process for collecting the expert demonstrations. 
The process consists of a $\Call{pathFollowExpert}{}$ used to transfer the dynamic object above the stationary object, similar to the original demonstrator.
The $\Call{insertExpert}{}$ is then used to precisely align the holes and pegs of the adapters to complete the task, using the features of the objects to create a feedback signal. 
The $\Call{insertExpert}{}$ is omitted for brevity in Algorithm \ref{alg:Expert}, but it can be used independently of the first stage (hence its independence of time $t$ and demonstration index $i$).

\subsection{Environments}
\label{sec:environments}
The base environment consists of an 18-dimensional action space and a 36-dimensional observation space. 
The action space, as previously described in Section \ref{sec:bm_expert}, consists of a delta-position ($\Delta \bm{\rho}$) and delta-rotation ($\Delta \bm{\phi}$) command for both end-effectors. 
The observation space is characterized by (1) the difference in the "expert's pose at the hover position above the stationary adapter" and the "current end-effector pose"; (2) the cube-root of the distance between the end-effector and respective near-side pin; and (3) the forces and torques acting upon the gripper sensors. 
This observation space is "duplicated" for both arms, so it can be viewed as having an 18-dimensional observation per arm.

At the start of every episode, the stationary adapter remains at a fixed location and the dynamic adapter is placed at a randomly chosen starting location based on the 200 original expert demonstrations. 
During the original expert demonstrations, this adapter is placed at a randomly generated position on the right side of the robot's workspace. 
An environment reward is available at every timestep, although it is not used by the learning algorithms. 
The environment reward, which we use to help measure an algorithm's success (Section \ref{sec:results}), is defined as follows:
\begin{equation}
    R(\bm{s}, t) = \sum_{j=1}^J \left[ e^{{\gamma(\bm{x}^j(t) - \bm{x}_{\ast}^j(t))}^2} \oplus e^{\lambda(d(\bm{\phi}^j(t), \bm{\phi}_{\ast}^j(t)))^2}\right] - \eta + \omega
\end{equation}
where, for DoF $j$: $\bm{x}^j(t) - \bm{x}_{\ast}^j(t)$ is the difference between the end-effector position and original expert's at time $t$; $d(\bm{\phi}^j(t), \bm{\phi}_{\ast}^j(t))$ is the axis part of the quaternion difference of the end-effector (computed using the multiplicative inverse); $\eta$ is a time penalty; $\omega$ is a positive reward for successfully inserting the adapter; and $\oplus$ is used to concatenate the six terms in the summand with the previous iteration, and then take mean of this result after iteration $J$ (for lack of better notation). 
There are three environments in total, namely, the Zero Noise, Low Noise, and High Noise environments (Section \ref{sec:act_obs_noise} for more detail). 
In all environments, $\gamma=-10$, $\lambda=-10$, $\eta = 1$, and $\omega = 100$, resulting in an average expert reward of $64.03 \pm 0.45$ (1 SE) over 600 demonstrations.

\section{Experimental Setup}
\label{sec:exp_setup}
The experimental procedure is divided into three phases: Analysis of Action/Observation Noise (\ref{sec:act_obs_noise}), Hyperparameter Search (\ref{sec:hp_search}), and Analysis of Hyperparameter Sensitivity (\ref{sec:sensitivity}). The results of each phase are used to provide an interpretation of key metrics presented in Section \ref{sec:results}.

\subsection{Action and Observation Noise Analysis}
\label{sec:act_obs_noise}

We study the effects of both observation noise and action noise on the success of the learning algorithms.
As such, a noise perturbation is applied to the actions ($\bm{a}'_i = m_{a_i}\bm{a}_i + b_{a_i}$) and observations ($\bm{o}'_{i} = m_{o_i}\bm{o}_i + b_{o_i}$) of the expert and environment, respectively.
Here, $i$ denotes a subgroup of the full vector (e.g., the change in the left end-effector position), and the time subscript is omitted for brevity.
The noisy actions, representing task space positions in this particular case, are not only used to augment the dataset but they are also passed to the inverse dynamics model. 
The bias terms, $b_{a_i}$ and $b_{o_i}$, are randomly sampled from the uniform distributions $\text{Unif}[b^{\text{min}}_{a_i}, b^{\text{max}}_{a_i}]$ and $\text{Unif}[b^{\text{min}}_{o_i}, b^{\text{max}}_{o_i}]$, respectively. 
The scaling terms, $m_{a_i}$ and $m_{o_i}$, are uniformly distributed about $1$, having endpoints $[0.9, 1.1]$ in the Zero Noise environment and $[0.7, 1.3]$ in  the High Noise environment. 
For rotation and delta-rotation features, the components are randomly rotated by $x_{a_i} \sim \text{Unif}[x^{\text{min}}_{a_i}, x^{\text{max}}_{a_i}]$, $y_{a_i} \sim \text{Unif}[y^{\text{min}}_{a_i}, y^{\text{max}}_{a_i}]$, and $z_{a_i} \sim \text{Unif}[z^{\text{min}}_{a_i}, z^{\text{max}}_{a_i}]$, constituting the role, pitch, and yaw perturbations, respectively. 
The same process applies to observation features. 
The widths of these uniform intervals are static properties defined in both the high-noise and low-noise environments, affecting the agent during training.

\begin{table}[!t]
    \begin{minipage}{.5\linewidth}        
       \vspace{3mm}
       \resizebox{\textwidth}{!}{
        \begin{tabular}{|l|l|}
        \hline \rowcolor{dagger} & \\ \noalign{\vspace{-2mm}}
        \rowcolor{dagger} \textbf{DAgger HP} & \textbf{HP Search Points} \\
        \hline & \\ \noalign{\vspace{-2mm}}
        \texttt{Learn Rate} & \texttt{[\textbf{5e-5}, 1e-4, 5e-4]} \\ 
        \texttt{$\pi$ Layers} & \texttt{[2, \textbf{3}]} \\ 
        \texttt{$\pi$ Units} & \texttt{[256, \textbf{512}]} \\ 
        \texttt{$\pi$ Activation} & \texttt{[Relu, \textbf{Tanh}]} \\ 
        \texttt{Normalize} & \texttt{[Expert, \textbf{None}]} \\ 
        \texttt{Epochs} & \texttt{[\textbf{64}, 128]} \\ 
        \texttt{Decay $\beta$} & \texttt{[\textbf{0.9}, 0.95]} \\ 
        \texttt{BC L1 $\lambda$} & \texttt{[0.0, \textbf{1e-6}, 1e-4]} \\
        \texttt{BC L2 $\lambda$} & \texttt{[0.0, \textbf{1e-6}, 1e-4]} \\
        \texttt{BC Batch Sz.} & \texttt{[128, \textbf{256}]} \\
        \hline
        \end{tabular}
        }
        \label{tab:dagger_hp}

        \vspace{2mm}

        \resizebox{\textwidth}{!}{
        \raggedright 
            \begin{tabular}{|l|l|}
            \hline \rowcolor{gail} & \\ \noalign{\vspace{-2mm}}
            \rowcolor{gail} \textbf{GAIL HP} & \textbf{HP Search Points} \\
            \hline & \newline \\ \noalign{\vspace{-2mm}}
            \texttt{$\pi$ Layers} & \texttt{[\textbf{2}, 3]} \\ 
            \texttt{$\pi$ Units} & \texttt{[\textbf{256}, 512]} \\ 
            \texttt{$\pi$ Activation} & \texttt{[Relu, \textbf{Tanh}]} \\ 
            \texttt{$\pi$ Max K.L} & \texttt{[\textbf{1e-2}, 3e-2]} \\
            \texttt{$\pi$ C.G Damping} & \texttt{[0.1, \textbf{0.3}]} \\ 
            \texttt{$\pi$ Ent. Reg} & \texttt{[0.0, \textbf{1e-3}, 1e-2]} \\
            \texttt{Normalize} & \texttt{[\textbf{Expert}, None]} \\ 
            \texttt{$\mathcal{R}$ Learn Rate} & \texttt{[1e-5, \textbf{5e-5}, 1e-4]} \\ 
            \texttt{$\mathcal{R}$ Layers} & \texttt{[1, \textbf{2}]} \\
            \texttt{$\mathcal{R}$ Units} & \texttt{[128, \textbf{256}]} \\
            \texttt{$\mathcal{R}$ Activation} & \texttt{[\textbf{Relu}, Tanh]} \\
            \texttt{$\mathcal{R}$ Ent. Reg.} & \texttt{[0.0, \textbf{1e-3}, 1e-2]} \\
            \texttt{Discount $\lambda$} & \texttt{[0.97, \textbf{0.99}]} \\
            \texttt{$\mathcal{V}$ Layers} & \texttt{[1, \textbf{2}]} \\ 
            \texttt{$\mathcal{V}$ Units} & \texttt{[\textbf{128}, 256]} \\
            \texttt{$\mathcal{V}$ Activation} & \texttt{[\textbf{Relu}, Tanh]} \\
            \texttt{$\mathcal{V}$ Max K.L} & \texttt{[\textbf{1e-2}, 3e-2]} \\
            \texttt{$\mathcal{V}$ C.G Damping} & \texttt{[\textbf{0.1}, 0.3]} \\
            \hline
            \end{tabular}
        }
        \label{tab:gail_hp}

        \vspace{2mm}

        \resizebox{\textwidth}{!}{
        \begin{tabular}{|l|l|}
        \hline \rowcolor{diffusion} & \\ \noalign{\vspace{-2mm}}
        \rowcolor{diffusion} \textbf{Diffusion HP} & \textbf{HP Search Points} \\
        \hline & \\ \noalign{\vspace{-2mm}}
        \texttt{Learn Rate} & \texttt{[5e-5, \textbf{1e-4}, 5e-4]} \\
        \texttt{Adam Decay} & \texttt{[\textbf{1e-6}, 1e-3]} \\
        \texttt{Batch Sz.} & \texttt{[128, 256, \textbf{512}]} \\
        \texttt{Diff. Steps} & \texttt{[50, \textbf{100}]} \\
        \texttt{L.R. Warmup} & \texttt{[\textbf{500}, 1000]} \\
        \hline
        \end{tabular}
        }
    
    \end{minipage}%
    \begin{minipage}{.5\linewidth}
        \resizebox{\textwidth}{!}{
        \begin{tabular}{|l|l|}
        \hline \rowcolor{bclone} & \\ \noalign{\vspace{-2mm}}
        \rowcolor{bclone} \textbf{BC HP} & \textbf{HP Search Points} \\
        \hline & \\ \noalign{\vspace{-2mm}}
        \texttt{Learn Rate} & \texttt{[5e-5, 1e-4, \textbf{5e-4}]} \\
        \texttt{$\pi$ Layers} & \texttt{[2, \textbf{3}]} \\
        \texttt{$\pi$ Units} & \texttt{[256, \textbf{512}]} \\ 
        \texttt{$\pi$ Activation} & \texttt{[Relu, \textbf{Tanh}]} \\
        \texttt{Normalize} & \texttt{[\textbf{Expert}, None]} \\
        \texttt{BC L1 $\lambda$} & \texttt{[0, 1e-6, \textbf{1e-4}]} \\
        \texttt{BC L2 $\lambda$} & \texttt{[0, 1e-6, \textbf{1e-4}]} \\
        \texttt{Batch Sz.} & \texttt{[\textbf{128}, 256, 512]} \\
        \hline
        \end{tabular}
        }

        \vspace{2mm}

        \resizebox{\textwidth}{!}{
        \begin{tabular}{|l|l|}
        \hline \rowcolor{ibc} & \\ \noalign{\vspace{-2mm}}
        \rowcolor{ibc} \textbf{IBC HP} & \textbf{HP Search Points} \\
        \hline & \\ \noalign{\vspace{-2mm}}
        \texttt{Learn Rate} & \texttt{[5e-5, \textbf{1e-4}, 5e-4]} \\ 
        \texttt{$\pi$ Layers} & \texttt{[2, \textbf{4}]} \\ 
        \texttt{$\pi$ Units} & \texttt{[\textbf{256}, 512]} \\
        \texttt{$\pi$ Activation} & \texttt{[\textbf{Relu}, Tanh]} \\
        \texttt{Dropout Rate} & \texttt{[\textbf{0.0}, 0.1, 0.2]} \\ 
        \texttt{Norm Batch Sz.} & \texttt{[50, \textbf{100}]} \\
        \texttt{Norm Samples} & \texttt{[1e3, \textbf{5e3}]} \\ 
        \texttt{Action Samples} & \texttt{[256, \textbf{512}, 1024]} \\
        \texttt{Pct. Langevin} & \texttt{[0.8, \textbf{1.0}]} \\
        \texttt{Langevin Iter.} & \texttt{[50, \textbf{100}]} \\
        \texttt{Counter Ex.} & \texttt{[8, 16, \textbf{32}]} \\
        \texttt{Batch Sz.} & \texttt{[256, \textbf{512}]} \\ 
        \texttt{Replay Sz.} & \texttt{[1e3, \textbf{1e4}]} \\
        \hline
        \end{tabular}
        }

        \vspace{2mm}

        \resizebox{\textwidth}{!}{
        \begin{tabular}{|l|l|}
        \hline \rowcolor{act} & \\ \noalign{\vspace{-2mm}}
        \rowcolor{act} \textbf{ACT HP} & \textbf{HP Search Points} \\
        \hline & \\ \noalign{\vspace{-2mm}}
        \texttt{Batch Sz.} & \texttt{[\textbf{256}, 512]} \\
        \texttt{Enc. Layers} & \texttt{[\textbf{1}, 2, 3]} \\
        \texttt{Dec. Layers} & \texttt{[\textbf{1}, 2, 3]} \\
        \texttt{Latent Dim} & \texttt{[\textbf{8}, 16]} \\
        \texttt{Attn. Heads} & \texttt{[\textbf{4}, 8]} \\
        \texttt{Learn Rate} & \texttt{[5e-5, \textbf{1e-4}, 5e-4]} \\
        \texttt{Dropout Rate} & \texttt{[0.0, \textbf{0.1}, 0.2]} \\
        \texttt{Hidden Dim} & \texttt{[\textbf{128}, 256]} \\
        \texttt{$\pi$ Units} & \texttt{[256, \textbf{512}]} \\
        \texttt{Activation} & \texttt{[\textbf{Relu}, Gelu]} \\
        \texttt{KL Weight} & \texttt{[1, 10, \textbf{100}]} \\
        \hline
        \end{tabular}
        }
        
    \end{minipage}
    \caption{Hyperparameter search for all algorithms. Bold denotes the value used for policy training.}

\label{tab:hp_search}
\end{table}

\subsection{Hyperparameter Search}
\label{sec:hp_search}
A hyperparameter search is conducted to obtain the best parameters for all algorithms, using the Zero Noise environment with 200 expert demonstrations. 
As a search over all such values is computationally infeasible, the Optuna \cite{akiba2019optuna} library is used to search over a discrete set of values for each hyperparameter.
For a given set of hyperparameters, 10 evenly spaced evaluations during training are conducted, and the evaluation resulting in the highest average reward (out of 10 rollouts) is returned to the optimizer.
For every algorithm, the optimizer iterates over many hyperparameter configurations to maximize this highest average reward.
Each algorithm's best (i.e., reward maximizing) hyperparameters are bolded in Table \ref{tab:hp_search}. 
We note that for behavioral cloning-based algorithms, alternative hyperparameter configurations achieved performance similar to that of the selected configuration. 
In the event that two configurations produce nearly identical results, we opted for the model with a simpler architecture to help prevent overfitting. 
We found that the default U-Net architecture for the Diffusion Policy (having channel sizes $\left[256,512,1024\right]$) worked well without the need for additional hyperparameter tuning. Additionally, using an action, observation, and prediction horizon of $4$, $4$, and $8$, respectively (introduced in Diffusion \cite{chi2023diffusion}) worked well for both ACT and Diffusion without the need for tuning.

\begin{table}[t]
    \vspace{2mm}
    \centering
    \resizebox{0.485\textwidth}{!}{%
    \begin{tabular}{
    l
    S[table-format=3]
    S[table-format=1.2, minimum-decimal-digits=2]
    @{${}\pm{}$}
    S[table-format=1.2, minimum-decimal-digits=2]
    S[table-format=1.2]
    @{${}\pm{}$}
    S[table-format=1.2, minimum-decimal-digits=2]
    S[table-format=1.2]
    @{${}\pm{}$}
    S[table-format=1.2, minimum-decimal-digits=2]
    }
    \hline \rowcolor{lightgray} &  & \multicolumn{2}{l}{} & \multicolumn{2}{l}{} & \multicolumn{2}{l}{} \\ \noalign{\vspace{-2mm}}
    \rowcolor{lightgray} &  & \multicolumn{2}{l}{\textbf{Zero Noise}} & \multicolumn{2}{l}{\textbf{Low Noise}} & \multicolumn{2}{l}{\textbf{High Noise}} \\
    \rowcolor{lightgray} {\textbf{Alg.}} & {\textbf{N.T.}} & \multicolumn{2}{l}{} & \multicolumn{2}{l}{} & \multicolumn{2}{l}{} \\
    \hline &  & \multicolumn{2}{l}{} & \multicolumn{2}{l}{} & \multicolumn{2}{l}{} \\ \noalign{\vspace{-2mm}}
    \multirow[t]{3}{*}{BC} & 50 & 0.36 & 0.01 & -1.06 & 0.13 & -2.43 & 0.13 \\
    & 100 & 0.4 & 0.0 & -0.23 & 0.05 & -1.16 & 0.08 \\
    & 200 & 0.43 & 0.0 & -0.05 & 0.03 & -0.55 & 0.04 \\

    \hline &  & \multicolumn{2}{l}{} & \multicolumn{2}{l}{} & \multicolumn{2}{l}{} \\ \noalign{\vspace{-2mm}}
    \multirow[t]{3}{*}{IBC} & 50 & 0.39 & 0.01 & -1.58 & 0.09 & -3.14 & 0.08 \\
    & 100 & 0.4 & 0.0 & -1.33 & 0.11 & -2.81 & 0.07 \\
    & 200 & 0.4 & 0.01 & -1.35 & 0.1 & -2.46 & 0.09 \\

    \hline &  & \multicolumn{2}{l}{} & \multicolumn{2}{l}{} & \multicolumn{2}{l}{} \\ \noalign{\vspace{-2mm}}
    \multirow[t]{3}{*}{DAgger} & {\textbf{\textemdash}} & 0.42 & 0.0 & 0.39 & 0.01 & 0.39 & 0.01 \\
    \hline &  & \multicolumn{2}{l}{} & \multicolumn{2}{l}{} & \multicolumn{2}{l}{} \\ \noalign{\vspace{-2mm}}
    \multirow[t]{3}{*}{GAIL} & 50 & 0.43 & 0.0 & 0.44 & 0.0 & 0.43 & 0.0 \\
    & 100 & 0.44 & 0.0 & 0.44 & 0.0 & 0.43 & 0.0 \\
    & 200 & 0.44 & 0.0 & 0.44 & 0.0 & 0.43 & 0.0 \\

    \hline &  & \multicolumn{2}{l}{} & \multicolumn{2}{l}{} & \multicolumn{2}{l}{} \\ \noalign{\vspace{-2mm}}
    \multirow[t]{3}{*}{ACT} & 50 & 0.43 & 0.0 & 0.43 & 0.0 & 0.43 & 0.0 \\
    & 100 & 0.42 & 0.0 & 0.42 & 0.0 & 0.42 & 0.0 \\
    & 200 & 0.42 & 0.0 & 0.42 & 0.0 & 0.42 & 0.0 \\

    \hline &  & \multicolumn{2}{l}{} & \multicolumn{2}{l}{} & \multicolumn{2}{l}{} \\ \noalign{\vspace{-2mm}}
    \multirow[t]{3}{*}{Diffusion} & 50 & 0.43 & 0.0 & 0.43 & 0.0 & 0.42 & 0.0 \\
    & 100 & 0.44 & 0.0 & 0.44 & 0.0 & 0.43 & 0.0 \\
    & 200 & 0.45 & 0.0 & 0.44 & 0.0 & 0.43 & 0.0 \\

    \hline &  & \multicolumn{2}{l}{} & \multicolumn{2}{l}{} & \multicolumn{2}{l}{} \\ \noalign{\vspace{-2mm}}
    \multirow[t]{3}{*}{Expert} & {\textbf{\textemdash}} & 0.40 & 0.0 & 0.40 & 0.0 & 0.40 & 0.0 \\
    \hline
    \end{tabular}
    }
    \captionsetup{justification=centering, skip=5pt}
    \caption{Action and Observation Noise Analysis-\\ Environment Reward (Normalized $\pm$ 1 SE)}
    \label{tab:noise_analysis}
\end{table}

\subsection{Hyperparameter Sensitivity Analysis}
\label{sec:sensitivity}
We conduct a sensitivity analysis on the hyperparameters of each algorithm to assess their local stability using the Zero Noise environment with 200 expert demonstrations. Local stability refers to an algorithm's robustness when hyperparameters deviate slightly from their optimal values. By perturbing each hyperparameter in both positive and negative directions, we form a hypercube, with edges representing extrema and the inner volume capturing all combinations of local interest. As the number of combinations grows exponentially with the number of hyperparameters, a grid search over all configurations is computationally infeasible. To address this, we use a NIST covering array \cite{forbes2008refining}, providing a computationally feasible subset of sampling configurations. Covering arrays have proven effective in tuning neural networks \cite{perez2016tuning}, and their systematic coverage is both effective and efficient \cite{algorain2023covering}. We perturb each hyperparameter's best value by $\pm$ 15\% and evaluate these combinations. For categorical hyperparameters, such as layer activation, we use the original values. For values requiring clipping, such as a discount factor, positive dropout rate, or DAgger $\beta$ decay, appropriate clipping is applied. This analysis assesses the algorithm's stability by evaluating its average performance under these perturbations.

\section{Results and Discussion}
\label{sec:results}

The results for the Action and Observation Noise Analysis (Section \ref{sec:act_obs_noise}) are shown in Table \ref{tab:noise_analysis}, and those for the Hyperparameter Sensitivity Analysis (Section \ref{sec:sensitivity}) are in Table \ref{tab:sensitivity_analysis}. Each cell in Table \ref{tab:noise_analysis} represents the average return from 10 different policies of the corresponding algorithm, run in the respective environment with random seed initialization. The "N.T" value limits the number of expert trajectories used to train the policy. For DAgger, which uses the oracle rather than demonstrations, expert trajectories are not utilized. During training, 10 evenly-spaced evaluations are conducted, each with 10 rollouts. The evaluation with the highest average return determines the policy's performance. This average return is standardized across all tagged rollouts from different algorithms. The same procedure applies to Table \ref{tab:sensitivity_analysis}.

\begin{table}[t]
    \vspace{2mm}
    \setlength{\tabcolsep}{4pt} 
    \renewcommand{\arraystretch}{1.2} 
    \centering
    \resizebox{0.485\textwidth}{!}{%
    \begin{tabular}{
    @{\hspace{7pt}}
    S[table-format=1.2]
    S[table-format=1.2]
    S[table-format=1.2]
    S[table-format=1.2]
    S[table-format=1.2]
    S[table-format=1.2]
    S[table-format=1.2]
    S[table-format=1.2]
    @{\hspace{2pt}}
    }
    \hline \rowcolor{lightgray} \multicolumn{1}{l}{} & \multicolumn{1}{l}{} & \multicolumn{1}{l}{} & \multicolumn{1}{l}{} & \multicolumn{1}{l}{} & \multicolumn{1}{l}{} & \multicolumn{1}{l}{} \\ \noalign{\vspace{-3mm}}
     \rowcolor{lightgray}  \multicolumn{1}{l}{\hspace{0mm} \textbf{BC}} & \multicolumn{1}{l}{\hspace{0mm} \textbf{IBC}} & \multicolumn{1}{l}{\hspace{0mm} \textbf{DAgger}} & \multicolumn{1}{l}{\hspace{0mm} \textbf{GAIL}} & \multicolumn{1}{l}{\hspace{0mm} \textbf{ACT}} & \multicolumn{1}{l}{\hspace{0mm} \textbf{Diffusion}} & \multicolumn{1}{l}{\hspace{0mm} \textbf{Expert}}\\
    \hline \multicolumn{1}{l}{} & \multicolumn{1}{l}{} & \multicolumn{1}{l}{} & \multicolumn{1}{l}{} & \multicolumn{1}{l}{} \\ \noalign{\vspace{-2mm}}
     0.35 & 0.25 & 0.30 & -1.35 & 0.30 & 0.33 & 0.29 \\
    \pm 0.00 & \pm 0.01 & \pm 0.00 & \pm 0.13 & \pm 0.00 & \pm 0.00 & \pm 0.00 \\
    \hline \multicolumn{1}{l}{} & \multicolumn{1}{l}{} & \multicolumn{1}{l}{} & \multicolumn{1}{l}{} & \multicolumn{1}{l}{} \\ \noalign{\vspace{-2mm}}
    \end{tabular}
    }
    \captionsetup{justification=centering, skip=0pt}
    \caption{Hyperparameter Sensitivity Analysis- \\ Environment Reward (Normalized $\pm$ 1 SE)}
    \label{tab:sensitivity_analysis}
\end{table}

Considering (a) maximizing reward is important across all algorithms and (b) different algorithms can produce their most successful policies at different times during training (e.g., due to overfitting or training instability), we find this method apt for comparing across all algorithms.
Consequently, the results should be interpreted carefully with the environment reward function in mind. 
For one, the policy's sequence of states during a rollout should ideally be close to the expert's.
The exponential reward (parameterized by the original expert's mean position and orientation at time $t$ for the same initial state) helps capture this objective. 
The reward also discourages policies that take too long to insert the adapter.
Most importantly, the reward largely accounts for success on the insertion task, as the environment only generates a large reward when the adapter is correctly inserted.

\begin{figure}[t]
    \centering
    \includegraphics[width=0.485\textwidth]{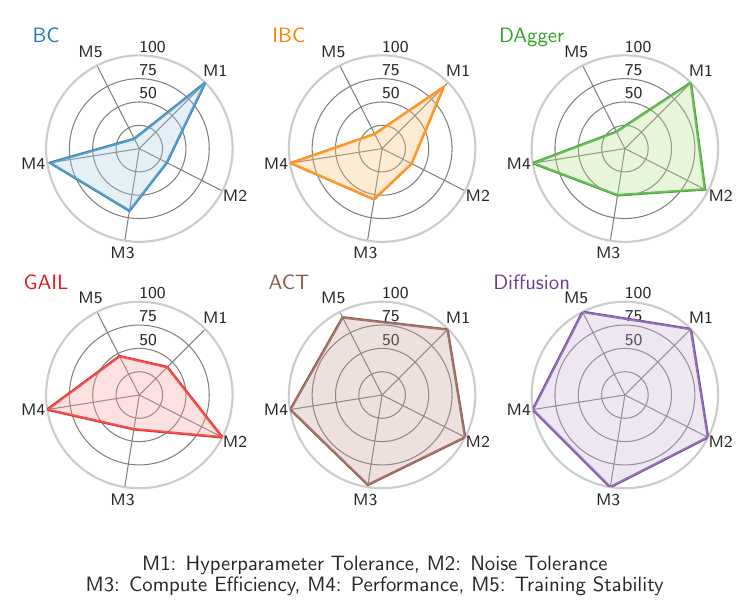}
    \caption{A high-level interpretation of the key metrics describing the algorithms in the bimanual insertion environments.}
    \label{fig:spider}
    \vspace{-1mm}
\end{figure}

Figure \ref{fig:spider} summarizes our findings from the experimental procedure (Section \ref{sec:exp_setup}). For the \textit{Hyperparameter Tolerance} metric (Section \ref{sec:hp_search}), we calculate the average percent success (where success indicates the adapter is correctly placed) across all hyperparameter configurations. The \textit{Noise Tolerance} metric (Section \ref{sec:act_obs_noise}) reflects the average percent success in Low and High Noise environments for all tagged rollouts. \textit{Compute Efficiency} is the negative $\log$ of the average total time (in seconds) to reach the selected policy, linearly scaled to $[0,100]$. We use the $\log$ scale due to the wide variance in runtime, exemplified by GAIL’s $128,000$ training rollouts per policy (approx. 2 days). \textit{Performance} indicates the average percent success in the Zero Noise environment across all numbers of expert demonstrations. \textit{Training Stability} is determined by the frequency of evaluation intervals (during training) where the average return shifts from positive to negative, suggesting a decline in policy success. A higher value implies more instability; thus, we use the negative average, scaled to $[0,100]$.

\begin{wrapfigure}[12]{r}{0.27\textwidth}
    \hspace{-4.5mm}
    \includegraphics[width=0.3\textwidth]{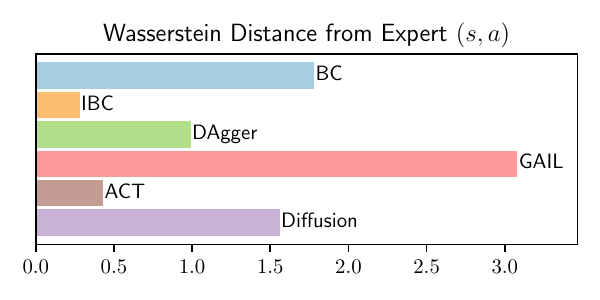}
    \captionsetup{skip=-3pt}
    \caption{The Wasserstein distance, as measured by the state-action samples from the best policies in the Zero Noise environment with 200 expert demonstrations.}
    \label{fig:wasserstein}
\end{wrapfigure}

\pagebreak

We also compare how close the learned state-action distributions of the policies are compared to the expert, using optimal transport. 
The Wasserstein distance is chosen due to its effectiveness in comparing samples from distributions that are not easily described in parametric form. 
For ACT and Diffusion, the most recent observation and the first action in the predicted sequence are paired together.
While there is no clear relationship between the Wasserstein distance and the algorithm's performance, we observe that Diffusion, GAIL, and BC -- which produce the highest average rewards in this scenario -- also have the largest Wasserstein distances from the expert.

We observe the following noteworthy properties. First, algorithms that either (a) interact with the environment (i.e., GAIL and DAgger) or (b) perform action/observation chunking (i.e., ACT and Diffusion) are more robust to noise perturbations. 
In the former case, this highlights the benefit of exploring during training and the effectiveness of having an oracle in the presence of increased noise-- at the expense of more computing and training time. 
In the latter case, we observe that the action and observation horizons introduced by Diffusion and ACT help cope with potentially non-Markovian environments. 
However, this design choice increases the dimensionality of the observation and action space (unless using, e.g., a recurrent model for dimensionality reduction), making it less suitable for RL-based methods and IBC, which we observe to be more susceptible to the curse of dimensionality.
As an ablation study, we found that performing observation and action chunking can improve the performance of BC in noisy environments. 
Furthermore, we observe that the time required to generate an action during evaluation is longer for IBC and Diffusion; however, for Diffusion, this time is largely affected by the number of de-noising iterations and the length of chunking horizons (i.e., the control frequency).

GAIL, Diffusion, and ACT obtained high rewards in all environments, indicating they are viable options for bimanual manipulation. 
However, GAIL performs the worst in the hyperparameter sensitivity category, and while an improvement would be made here with an increased maximum time limit for training (approx. 2 days), other algorithms can achieve suitable results in a matter of hours or less. 
These results indicate that ACT and Diffusion are favorable options, whereas GAIL -- and to some extent, DAgger -- are suitable options for tasks that benefit from extensive interaction with the environment during training. 
Furthermore, our findings regarding training stability and high performance in fine-grained manipulation environments align with those reported for Diffusion Policy \cite{chi2023diffusion} and ACT \cite{zhao2023learning}, respectively.

\section{Conclusion}

We present a comprehensive assessment of various imitation learning algorithms in a bimanual manipulation environment.
A carefully selected set of experiments is conducted to evaluate the key characteristics inherent to these algorithms, such as sample efficiency, sensitivity to perturbations in hyperparameter values, and robustness under observation and action noise-- to name a few.  
This investigation leads to new insights regarding the applicability of imitation learning for fine-grained industrial tasks and highlights the effectiveness of the chosen methodology. 
Furthermore, we discuss the implications of selecting these approaches, elucidating how they behave at an empirical and conceptual level.
While the presented approaches can learn to complete a complex sequential insertion task, further investigation into deploying these methods in physical environments will help glean insights into the feasibility of using these systems in everyday situations.

\end{document}